\documentclass[runningheads,a4paper]{llncs}

\usepackage{amssymb}
\usepackage{graphicx}

\usepackage{amsmath}
\usepackage{siunitx}
\usepackage{graphicx}
\usepackage{booktabs}
\usepackage{caption}
\usepackage{authblk} 

\usepackage{algorithm}
\usepackage[noend]{algpseudocode}
\makeatletter
\def\BState{\State\hskip-\ALG@thistlm}
\makeatother

\def\onedot{. }
\def\eg{\emph{e.g., }} 
\def\ie{\emph{i.e., }}

\def\wrt{w.r.t\onedot}

\usepackage{url}

\begin{document}

\mainmatter  

\title{Simultaneously Learning Architectures and Features of Deep Neural Networks}

\titlerunning{Simul. Learning Arch. and Feat. of Deep Neural Networks}

\author{%
	Tinghuai Wang\inst{1} \and 
	Lixin Fan\inst{1} \and 
	Huiling Wang\inst{2}
}%
\institute{
	Nokia Technologies, Finland
	\and
	Tampere University, Finland}

%
%
%


%
%

\maketitle

\begin{abstract}
This paper presents a novel method which simultaneously learns the \textit{number of filters} and \textit{network features} repeatedly over multiple epochs. We propose a novel pruning loss to explicitly enforces the optimizer to focus on promising candidate filters while suppressing contributions of less relevant ones. In the meanwhile, we further propose to enforce the diversities between filters and this diversity-based regularization term improves the trade-off between model sizes and accuracies. It turns out the interplay between architecture and feature optimizations improves the final compressed models, and the proposed method is compared favorably to existing methods, in terms of both models sizes and accuracies for a wide range of applications including image classification, image compression and audio classification.

\end{abstract}

\section{INTRODUCTION}


Large and deep neural networks, despite of their great successes in a wide variety of applications, call for compact and efficient model representations to reduce the vast amount of network parameters and computational operations, that are resource-hungry in terms of memory, energy and communication bandwidth consumption.  This need is imperative especially for resource constrained devices such as mobile phones, wearable and Internet of Things (IoT) devices. Neural network compression is a set of techniques that address these challenges raised in real life industrial applications. 

Minimizing network sizes without compromising original network performances has been pursued by a wealth of methods, which often adopt a three-phase learning process, i.e. training-pruning-tuning. In essence, network features are first learned, followed by the pruning stage to reduce network sizes.  The subsequent fine-tuning phase aims to restore deteriorated performances incurred by undue pruning. This ad hoc three phase approach, although empirically justified e.g. in \cite{NetSlim_Liu2017d,IntreprePrune_Qin2018,Lib,Wen2016,Zhou2016a}, was recently questioned with regards to its efficiency and effectiveness. Specifically \cite{RethinkPrune_2018arXiv,PruningNip_2018arXiv} argued that the network \textit{architecture} should be optimized first, and then \textit{features} should be learned from scratch in subsequent steps.   

In contrast to the two aforementioned opposing approaches, the present paper illustrates a novel method which simultaneously learns 
both the \textit{number of filters} and \textit{network features}  over multiple optimization epochs. This integrated optimization process brings about immediate benefits and challenges --- on the one hand, separated processing steps such as training, pruning, fine-tuning etc, are no longer needed and the integrated optimization step guarantees consistent performances for the given neural network compression scenarios. On the other hand,  the dynamic change of network architectures has significant influences on the optimization of features, which in turn might affect the optimal network architectures. It turns out the interplay between architecture and feature optimizations plays a crucial role in improving the final compressed models. 

\section{RELATED WORK}

Network pruning was pioneered \cite{lecun1990optimal,hassibi1993second,han2015learning} in the early development of neural network, since when a broad range of methods have been developed. 
We focus on neural network compression methods that prune filters or channels. For thorough review of other approaches we refer to a recent survey paper \cite{Cheng2017}. 


Li \emph{et al}. \cite{Lib} proposed to prune filters with small effects on the output accuracy and managed to reduce  about one third of inference cost without compromising original accuracy on CIFAR-10 dataset. Wen \emph{et al}. \cite{Wen2016} proposed a structured sparsity regularization framework, in which the group lasso constrain term was incorporated to penalize and remove unimportant filters and channels.  Zhou \emph{et al}. \cite{Zhou2016a} also adopted a similar regularization framework, with tensor trace norm and group sparsity incorporated to penalize the number of neurons.  Up to 70\% of model parameters were reduced without scarifying classification accuracies on CIFAR-10 datasets. Recently Liu \emph{et al}. \cite{NetSlim_Liu2017d} proposed an interesting network slimming method, which imposes L1 regularization on channel-wise \textit{scaling factors} in batch-normalization layers and demonstrated remarkable compression ratio and speedup using a surprisingly simple implementation. Nevertheless, network slimming based on scaling factors is not guaranteed to achieve desired accuracies and separate fine-tunings are needed to restore reduced accuracies. 
Qin \emph{et al}. \cite{IntreprePrune_Qin2018}  proposed a functionality-oriented filter pruning method to remove less important filters, in terms of their contributions to classification accuracies. It was shown that the efforts for model retraining is moderate but still necessary, as in the most of state-of-the-art compression methods. 

DIVNET adopted Determinantal Point Process (DPP) to enforce diversities between individual neural activations \cite{DivNet_2015arXiv}. 
Diversity of filter weights defined in (\ref{eq:diversity_ncc}) is related to orthogonality of weight matrix, which has been extensively studied. An example being \cite{Harandi2016}, proposed to learn Stiefel layers, which have orthogonal weights, and demonstrated its applicability in compressing network parameters.  Interestingly, the notion of diversity regularized machine (DRM) has been proposed to generate an ensemble of SVMs in the PAC learning framework \cite{Yu2011}, yet its definition of diversity is critically different from our definition in (\ref{eq:diversity_ncc}), and its applicability to deep neural networks is unclear. 

\section{SIMULTANEOUS LEARNING OF ARCHITECTURE AND FEATURE}

The proposed compression method belongs to the general category of filter-pruning approaches. 
In contrast to existing methods \cite{NetSlim_Liu2017d,IntreprePrune_Qin2018,Lib,Wen2016,Zhou2016a,RethinkPrune_2018arXiv,PruningNip_2018arXiv}, we adopt following techniques to ensure that simultaneous optimization of network architectures and features is a technically sound approach. First, we introduce an explicit \textit{pruning loss} estimation as an additional regularization term in the optimization objective function.  As demonstrated by experiment results in Section \ref{sect:exper}, the introduced pruning loss enforces the optimizer to focus on promising candidate filters while suppressing contributions of less relevant ones.  Second, based on the importance of filters,  we explicitly \textit{turn-off} unimportant filters below given percentile threshold.  We found the explicit shutting down of less relevant filters is indispensable to prevent biased estimation of pruning loss.  Third, we also propose to enforce the diversities between filters and this diversity-based regularization term improves the trade-off between model sizes and accuracies, as demonstrated in various applications.  

Our proposed method is inspired by network slimming \cite{NetSlim_Liu2017d} and main differences from this prior art are two-folds: a) we introduce the pruning loss and incorporate explicit pruning into the learning process, without resorting to the multi-pass pruning-retraining cycles; b) we also introduce filter-diversity based regularization term which improves the trade-off between model sizes and accuracies. 

\subsection{Loss Function}\label{sect:review-sparsity}

Liu \emph{et al}. \cite{NetSlim_Liu2017d} proposed to push towards zero the scaling factor in batch normalization (BN) step during learning, and subsequently, the insignificant channels with small scaling factors are pruned. This sparsity-induced penalty is introduced by regularizing L1-norm of the learnable parameter $\gamma$ in the BN step \ie
\begin{equation}\label{eq:filter_spartiy}
g(\gamma) = \left| \gamma \right|; \textnormal{ where }  \hat{z}= \frac{z_{in} - \mu_B}{\sqrt{\sigma^2 + \epsilon}};  z_{out} = \gamma \hat{z} + \beta,
\end{equation}
in which $z_{in}$ denote filter inputs, $\mu_B, \sigma$ the filter-wise mean and variance of inputs, $\gamma, \beta$ the scaling and offset parameters of batch normalization (BN) and $\epsilon$ a small constant to prevent numerical un-stability for small variance.  It is assumed that there is always a BN filter appended after each convolution and fully connected filter, so that the scaling factor $\gamma$ is directly leveraged to prune unimportant filters with small $\gamma$ values. Alternatively, we propose to directly introduce scaling factor to each filter since it is more universal than reusing BN parameters, especially considering the networks which have no BN layers.

By incorporating a filter-wise sparsity term, the object function to be minimized is given by: 
\begin{equation}\label{eq:regularized_sparisty_func}
L = \sum_{(\textbf{x},y)} loss( f(\textbf{x},\textbf{W}), y)  + \lambda \sum_{\gamma \in \Gamma } g(\gamma),
\end{equation}
where the first term is the task-based loss, $g(\gamma)=||\gamma||_1$ and $\Gamma$ denotes the set of scaling factors for all filters. This pruning scheme, however, suffers from two main drawbacks: 1) since scaling factors are equally minimized for all filterers, it is likely that the pruned filters have unignorable contributions that should not be unduly removed.
2) the pruning process, \ie architecture selection, is performed independantly \wrt the feature learning; the performance of pruned network is inevitably compromised and has to be recovered by single-pass or multi-pass fine-tuning, which impose additional computational burdens. 

\subsubsection{An integrated optimization}\hfill

Let $\textbf{W}, \check{\textbf{W}}, \hat{\textbf{W}}$ denote the sets of neural network weights for, respectively, all filters, those pruned and remained ones i.e. $\textbf{W} = \{ \check{\textbf{W}} \bigcup \hat{\textbf{W}} \}$.   In the same vein, $\Gamma= \{ P(\Gamma) \bigcup R(\Gamma)\}$ denote the sets of scaling factors for all filters, those removed and remained ones respectively.

To mitigate the aforementioned drawbacks, we propose to introduce two additional regularization terms to Eq. \ref{eq:regularized_sparisty_func},
\begin{align}\label{eq:regularized_all} 
L( \hat{\textbf{W}}, R(\Gamma)) = & \sum_{(\textbf{x},y)} loss( f(\textbf{x},\hat{\textbf{W}} ), y) + \lambda_1  \sum_{\gamma \in R(\Gamma) } g(\gamma) \nonumber \\ 
&  - \lambda_2 \frac{\sum_{\gamma \in R({\Gamma}) } \gamma }{\sum_{\gamma \in \Gamma } \gamma} - \lambda_3 \sum_{l \in L} Div(\hat{\textbf{W}}^l),
\end{align}
where $loss( \cdot, \cdot )$ and $\sum_{\gamma \in R(\Gamma) } g(\gamma)$ are defined as in Eq. \ref{eq:regularized_sparisty_func}, the third term is the pruning loss  and the forth is the diversity loss which are elaborated below. $\lambda_1, \lambda_2, \lambda_3$ are  weights of corresponding regularization terms. 

\begin{figure*}[t!]
	\centering
	\includegraphics[width=.45\textwidth]{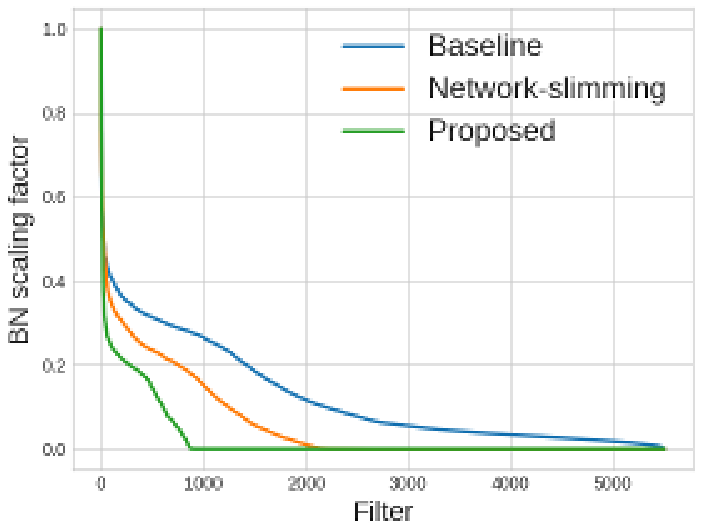}
	\includegraphics[width=.45\textwidth]{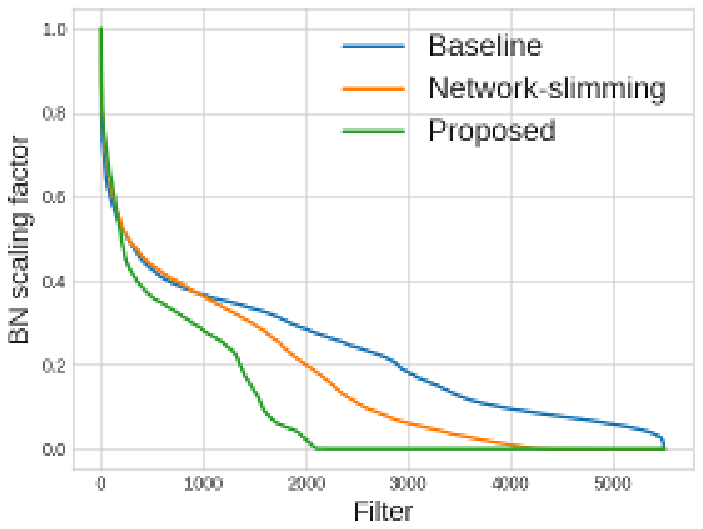}
	\caption{Comparison of scaling factors for three methods, \ie baseline with no regularization, network-slimming \cite{NetSlim_Liu2017d}, and the proposed method with diversified filters, trained with CIFAR-10 and CIFAR-100. 
		Note that the pruning loss defined in (\ref{eq:regularized_all}) are 0.2994, 0.0288, \num{1.3628e-6}, respectively, for three methods. Accuracy deterioration are 60.76\% and 0\% for network-slimming \cite{NetSlim_Liu2017d} and the proposed methods, and the baseline networks completely failed after pruning, due to insufficient preserved filters at certain layers.
	} \label{fig:nsf_compare}
\end{figure*}

\subsubsection{Estimation of pruning loss}\hfill

The second regularization term in (\ref{eq:regularized_all}) i.e. $\gamma^R := \frac{\sum_{\gamma \in R({\Gamma}) } \gamma }{\sum_{\gamma \in \Gamma } \gamma}$ (and its compliment $\gamma^P :=\frac{\sum_{\gamma \in P({\Gamma}) } \gamma }{\sum_{\gamma \in \Gamma } \gamma} = 1 - \gamma^R$)  is closely related to performance deterioration incurred by undue pruning\footnote{In the rest of the paper we refer to it as the estimated pruning loss.}.  
The scaling factors of pruned filters $ P(\Gamma)$, as in \cite{NetSlim_Liu2017d},  are determined by first ranking all $\gamma$ and taking those below the given percentile threshold. Incorporating this pruning loss  enforces the optimizer to increase scaling factors of promising filters while suppressing contributions of less relevant ones. 

The rationale of this pruning strategy can also be empirically justified in Figure \ref{fig:nsf_compare}, 
in which scaling factors of three different methods are illustrated. When the proposed regularization terms are added, clearly, we observed a tendency for scaling factors being dominated by few number of filters --- when 70\% of filters are pruned from a VGG network trained with CIFAR-10 dataset, the estimated pruning loss $ \frac{\sum_{\gamma \in P({\Gamma}) } \gamma }{\sum_{\gamma \in \Gamma } \gamma} $ equals  0.2994, 0.0288, \num{1.3628e-6}, respectively, for three compared methods. Corresponding accuracy deterioration are 60.76\% and 0\% for network-slimming \cite{NetSlim_Liu2017d} and the proposed methods. Therefore, retraining of pruned network is no longer needed for the proposed method, while \cite{NetSlim_Liu2017d} has to retain the original accuracy through single-pass or multi-pass of pruning-retraining cycles. 

\subsubsection{Turning off candidate filters}\hfill

It must be noted that the original loss $\sum_{(\textbf{x},y)} loss( f(\textbf{x},{\textbf{W}} ), y)$ is independent of the pruning operation.  If we adopt this loss in (\ref{eq:regularized_all}), the estimated pruning loss might be seriously biased because of undue assignments of $\gamma$ not being penalized. It seems likely some candidate filters are assigned with rather small scaling factors, nevertheless, they still retain decisive contributions to the final classifications. Pruning these filters blindly leads to serious performance deterioration, according to the empirical study, where we observe over 50$\%$ accuracy loss at high pruning ratio. 


In order to prevent such biased pruning loss estimation, we therefore explicitly shutdown the outputs of selected filters by setting corresponding scaling factors to absolute zero.  The adopted loss function becomes $\sum_{(\textbf{x},y)} loss( f(\textbf{x},\hat{\textbf{W}} ), y)$.
This way, the undue loss due to the biased estimation is reflected in $loss( f(\textbf{x},\hat{\textbf{W}}), y)$, which is minimized during the learning process.  We found the turning-off of candidate filters is indispensable.

\begin{algorithm}[t!]
	\caption{Proposed algorithm}\label{algo}
	\begin{algorithmic}[1]
		\Procedure{Online Pruning}{}
		\State $\textit{Training data}~ \gets \{x_i, y_i\}_{i=1}^{N} $
		\State $\textit{Target pruning ratio}~\mathbf{Pr}_N \gets \mathbf{p}\% $
		\State $\textit{Initial network weights}~W \gets \textit{method by \cite{he2015delving}}$
		\State $\Gamma \gets \{0.5\}$
		\State $\hat{W} \gets W$
		\State $P(\Gamma) \gets  \emptyset$
		\State $R(\Gamma) \gets  \Gamma$
		\For{\text{each} \textit{epoch $n \in$\{$1,\dots,N$\}}}
		\State $\textit{Current pruning ratio}~\mathbf{Pr}_n \in [0, \mathbf{Pr}_N]$
		\State $\textit{Sort}~\Gamma$
		\State $P(\Gamma) \gets \textit{prune filters \wrt} \mathbf{Pr}_n$
		\State $R(\Gamma) \gets \Gamma \setminus \textit{P} (\Gamma)$
		\State $\textit{Compute}~L( \hat{\textbf{W}}, R(\Gamma))~\textit{in Eq.}~(\ref{eq:regularized_sparisty_func})$
		\State $\hat{\textbf{W}} \gets \textit{SGD}$
		\State $\check{\textbf{W}} \gets \hat{\textbf{W}} \setminus \check{\textbf{W}}$
		\EndFor
		\EndProcedure
	\end{algorithmic}
\end{algorithm}

\subsubsection{Online pruning}

We take a global threshold for pruning which is determined by percentile among all channel scaling factors. The pruning process is performed over the whole training process, \ie simultaneous pruning and learning. To this end, we compute a linearly increasing pruning ratio from the first epoch (e.g., 0\%) to the last epoch (e.g., 100\%) where the ultimate pruning target ratio is applied. Such an approach endows neurons with sufficient evolutions driven by diversity term and pruning loss, to avoid mis-pruning neurons prematurely which produces crucial features. Consequently our architecture learning is seamlessly integrated with feature learning. After each pruning operation, a narrower and more compact network is obtained and its corresponding weights are copied from the previous network.



\subsubsection{Filter-wise diversity}\hfill

The third regularization term in (\ref{eq:regularized_all}) 
encourages high diversities between filter weights as shown below.    Empirically, we found that this term improves the trade-off between model sizes and accuracies (see experiment results in Section \ref{sect:exper}).  

We treat each filter weight, at layer $l$, as a weight (feature) vector $\textbf{W}^l_i$ of length $w \times h \times c$, where $w,h$ are filter width and height, $c$ the number of channels in the filter. The \textit{diversity} between two weight vectors of the same length is based on the \textit{normalized cross-correlation} of two vectors: 
\begin{align}\label{eq:diversity_ncc}
div(\textbf{W}_i, \textbf{W}_j) := 1 - |  \langle \mathbf{\bar{W}}_i, \mathbf{\bar{W}}_j  \rangle   | ,
\end{align}
in which $ \mathbf{\bar{W}} =  \frac{ \mathbf{{W}}}{ | \mathbf{{W}} | }  $ are normalized weight vectors, and $\langle \cdot , \cdot \rangle$  is the dot product of two vectors. Clearly, the diversity is bounded $0 \leq div(\textbf{W}_i, \textbf{W}_j)  \leq 1$, with value close 0 indicating low diversity between highly correlated vectors and values near 1 meaning high diversity between uncorrelated vectors. In particular, diversity equals 1 also means that two vectors are orthogonal with each other.

The \textit{diversities} between $N$ filters at the same layer $l$ is thus characterized by a \textit{N-by-N} matrix
in which elements $d_{ij} =div(\textbf{W}^l_i, \textbf{W}^l_j), i,j=\{1,\cdots,N\}$ are pairwise diversities between weight vectors $\textbf{W}^l_i, \textbf{W}^l_j$.  Note that for diagonal elements $d_{ii}$ are constant 0. The \textit{total diversity} between all filters is thus defined as the sum of all elements 
\begin{align}
Div(\textbf{W}^l) := \sum^{N,N}_{i,j=1,1} d_{i,j}.
\end{align}

\begin{table*}[t!]
	\caption{Results on CIFAR-10 dataset}
	\centering 
	\begin{tabular}{lccccc} 
		\toprule
		Models /\ Pruning Ratio & 0.0 &  0.5 & 0.6 & 0.7 & 0.8 \\
		\midrule
		VGG-19 (Base-line) & 0.9366 & - & - & - & - \\
		VGG-19 (Network-slimming) & -  & - & - & 0.9380 & NA \\
		VGG-19 (Ours) & - &  0.9353 & 0.9394 & 0.9393 & 0.9302 \\
		\midrule
		ResNet-164 (Base-line) & 0.9458 & -& - & - & - \\
		ResNet-164 (Network-slimming) & - &  - & 0.9473 & NA & NA \\
		ResNet-164 (Ours) & - &  0.9478 & 0.9483 & 0.9401 & NA \\
		\bottomrule
	\end{tabular} \label{tbl:c10} 
\end{table*}

\begin{table*}[t!]
	\caption{Results on CIFAR-100 dataset}
	\centering 
	\begin{tabular}{lccccc} 
		\toprule
		Models /\ Pruning Ratio & 0.0 &  0.3  & 0.4 & 0.5 & 0.6 \\
		\midrule
		VGG-19 (Base-line) & 0.7326 &  - & - & - & - \\
		VGG-19 (Network-slimming) & - & - & 0.7348 & - & -  \\
		VGG-19 (Ours) & - & 0.7332 & 0.7435 & 0.7340 & 0.7374  \\
		\midrule
		ResNet-164 (Base-line) & 0.7663 & - & - & - & - \\
		ResNet-164 (Network-slimming) & - & - & 0.7713 & - & 0.7609  \\
		ResNet-164 (Ours) & - & 0.7716 & 0.7749 & 0.7727 & 0.7745  \\
		\bottomrule
	\end{tabular} \label{tbl:c100} 
\end{table*}

\section{EXPERIMENT RESULTS} \label{sect:exper}
In this section, we evaluate the effectiveness of our method on various applications with both visual and audio data. 

\subsection{Datasets}
For visual tasks, we adopt ImageNet and CIFAR datasets. The ImageNet dataset contains 1.2 million training images and 50,000 validation images of 1000 classes. CIFAR-10 \cite{krizhevsky2009learning} which consists of 50K training and 10K testing RGB images with 10 classes. CIFAR-100 is similar to CIFAR-10, except it has 100 classes. The input image is 32$\times$32 randomly cropped from a zero-padded 40$\times$40 image or its flipping. For audio task, we adopt ISMIR Genre dataset  \cite{cano2006ismir} which has been assembled for training and development in the ISMIR 2004 Genre Classification contest. It contains 1458 full length audio recordings from Magnatune.com distributed across the 6 genre classes: Classical, Electronic, JazzBlues, MetalPunk, RockPop, World. 

\subsection{Image Classification}

We evaluate the performance of our proposed method for image classification on CIFAR-10/100 and ImageNet. We investigate both classical plain network, VGG-Net \cite{simonyan2014very}, and deep residual network \ie ResNet \cite{he2016deep}. We evaluate our method on two popular network architecture \ie VGG-Net \cite{simonyan2014very}, and ResNet \cite{he2016deep}. We take  variations of the original VGG-Net, \ie VGG-19 used in  \cite{NetSlim_Liu2017d} for comparison purpose. ResNet-164 which has 164-layer pre-activation ResNet with bottleneck structure is adopted. As base-line networks, we compare with the original networks without  regularization terms and  their counterparts in network-slimming \cite{NetSlim_Liu2017d}. For ImageNet, we adopt VGG-16 and ResNet-50 in order to compare with the original networks. 

To make a fair comparison with \cite{NetSlim_Liu2017d}, we adopt BN based scaling factors for optimization and pruning.  On CIFAR, we train all the networks from scratch using SGD with mini-batch size 64 for 160 epochs. The learning rate is initially set to 0.1 which is reduced twice by 10 at 50\% and 75\% respectively. Nesterov momentum \cite{sutskever2013importance} of 0.9 without dampening and a weight decay of $10^{-4}$ are used. The robust weight initialization method proposed by \cite{he2015delving} is adopted. We use the same channel sparse regularization term and its hyperparameter $\lambda = 10^{-4}$ as defined in \cite{NetSlim_Liu2017d}.

\begin{table*}[t!]
	\centering
		\caption{Accuracies of different methods before (orig.) and after pruning (pruned). For CIFAR10 and CIFAR100, 70\% and 50\% filters are pruned respectively. Note that 'NA' indicates the baseline networks completely failed after pruning, due to insufficient preserved filters at certain layers. }
	\begin{tabular}{|c|c|c|c|}
		\hline
		CIFAR10   & \multicolumn{3}{c|}{Methods}   \\
		& BASE & SLIM\cite{NetSlim_Liu2017d} & OURS \\		\hline
		\hline
		ACC orig. &  0.9377 & 0.9330  & 0.9388 \\
		\hline
		ACC pruned &  NA & 0.3254 & 0.9389 \\
		\hline
		$\gamma^P$ &  0.2994 & 0.0288  & 1.36e-6 \\
		\hline
	\end{tabular}
	\begin{tabular}{|c|c|c|c|}
		\hline
		CIFAR100   & \multicolumn{3}{c|}{Methods}   \\
		& BASE & SLIM\cite{NetSlim_Liu2017d} & OURS \\		\hline
		\hline
		ACC orig. & 0.7212 & 0.7205  & 0.75 \\
		\hline
		ACC pruned &  NA & 0.0531 & 0.7436 \\
		\hline
		$\gamma^P$ &  0.2224 & 0.0569  & {4.75e-4} \\
		\hline
	\end{tabular}
	\label{tab_acc_comp}
\end{table*}

\subsubsection{Overall performance}

The results on CIFAR-10 and CIFAR-100 are shown in Table \ref{tbl:c10} and Table \ref{tbl:c100} respectively. On both datasets, we can observe when typically 50-70\% fitlers of the evaluated networks are pruned, the new networks can still achieve accuracy higher than the original network. For instance, with 70\% filters pruned VGG-19 achieves an accuracy of 0.9393, compared to 0.9366 of the original model on CIFAR-10. We attribute this improvement to the introduced diversities between filter weights, which naturally provides discriminative feature representations in intermediate layers of networks. 

As a comparison, our method consistently outperforms network-slimming without resorting to fine-tuning or multi-pass pruning-retraining cycles. It is also worth-noting that our method is capable of pruning networks with prohibitively high ratios which are not possible in network-slimming. Take VGG-19 network on CIFAR-10 dataset as an example, network-slimming prunes as much as 70\%, beyond which point the network cannot be reconstructed as some layers are totally destructed. On the contrary, our method is able to reconstruct a very narrower network by pruning 80\% filters while producing a marginally degrading accuracy of 0.9302. We conjecture this improvement is enabled by our simultaneous feature and architecture learning which can avoid pruning filters prematurely as in network-slimming where the pruning operation (architecture selection) is isolated from the feature learning process and the performance of the pruned network can be only be restored via fine-tuning. 

The results on ImageNet are shown in Table \ref{tbl:imagenet} where we also present comparison with \cite{DataDrivenSS_Huang2018} which reported top-1 and top-5 errors on ImageNet. On VGG-16, our method provides 1.2\% less accuracy loss while saving additionally 20.5M parameters and 0.8B FLOPs compared with \cite{DataDrivenSS_Huang2018}. On ResNet-50, our method saves 5M more parameters and 1.4B more FLOPs than \cite{DataDrivenSS_Huang2018} while providing 0.21\% higher accuracy.

\begin{table*}[t!]
	\caption{Results on ImageNet dataset}
	\centering 
	\begin{tabular}{lccccccc} 
		\toprule
		Models & Top-1 & Top-5 &  Params &   FLOPs \\
		\midrule
		VGG-16 \cite{DataDrivenSS_Huang2018}&  31.47 & 11.8 & 130.5M & 7.66B \\
		VGG-16 (Ours) &  30.29  &  10.62& 44M & 6.86B  \\
		VGG-16 (Ours) &  31.51  &  11.92& 23.5M & 5.07B  \\
		\midrule
		ResNet-50 \cite{DataDrivenSS_Huang2018}  &  25.82 &  8.09 &  18.6M & 2.8B  \\
		ResNet-50 (Ours) &  25.61 &  7.91 &  13.6M & 1.4B  \\
		ResNet-50 (Ours) &  26.32 &  8.35 &  11.2M & 1.1B  \\
		\bottomrule
	\end{tabular} \label{tbl:imagenet} 
\end{table*}

\subsubsection{Ablation study}

In this section we investigate the contribution of each proposed component through ablation study. 

\paragraph{Filter Diversity}

\begin{figure*}[t!]
	\centering
	\includegraphics[width=.49\textwidth]{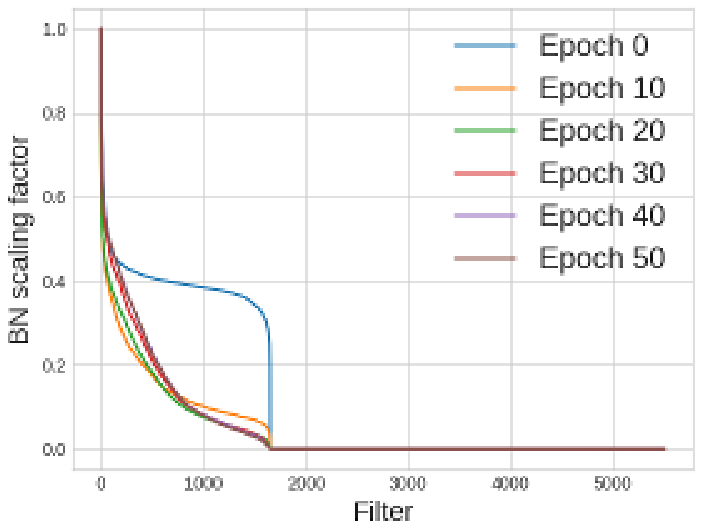} 
	\includegraphics[width=.49\textwidth]{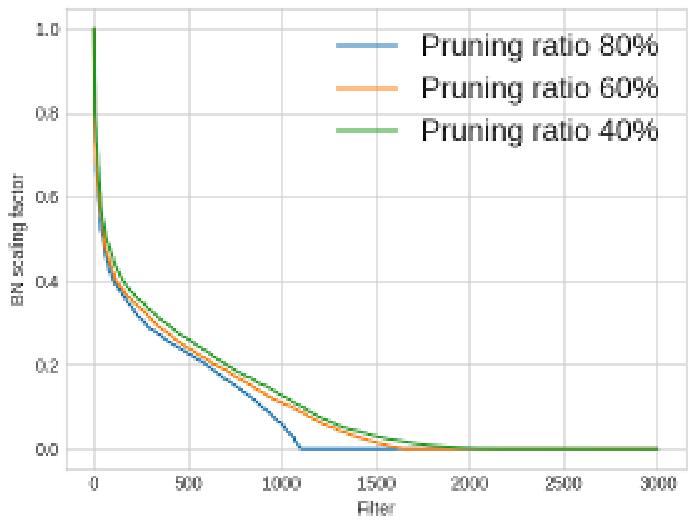} 
	\caption{(a) Scaling factors of the VGG-19 network at various epochs during training trained with diversified filters (b) Sorted scaling factors of VGG-19 network trained with various pruning ratios on CIFAR-10. 
	}\label{fig:scalings}
\end{figure*}

Fig. \ref{fig:scalings} (a) shows the sorted scaling factors of VGG-19 network trained with the proposed filter diversity loss at various training epochs. With the progress of training, the scaling factors become increasingly sparse and the number of large scaling factors, \ie the area under the curve, is decreasing.  Fig. \ref{fig:nsf_compare} shows the sorted scaling factors of VGG-19 network for the baseline model with no regularization, network-slimming \cite{NetSlim_Liu2017d}, and the proposed method with diversified filters, trained with CIFAR-10 and CIFAR-100. We observe significantly improved sparsity by introducing filter diversity to the network compared with network-slimming, indicated by \textit{nsf}. Remember the scaling factors essentially determine the importance of filters, thus, maximizing \textit{nsf} ensures that the deterioration due to filter pruning is minimized. Furthermore, 
the number of filters associated with large scaling factor is largely reduced, rendering more irrelevant filter to be pruned harmlessly. This observation is quantitatively confirmed in Table \ref{tab_acc_comp} which lists the accuracies of three schemes before and after pruning for both CIFAR-10 and CIFAR-100 datasets.   It is observed that retraining of pruned network is no longer needed for the proposed method, while network-slimming has to restore the original accuracy through single-pass or multi-pass of pruning-retraining cycles. Accuracy deterioration are 60.76\% and 0\% for network-slimming and the proposed method respectively, whilst the baseline networks completely fails after pruning, due to insufficient preserved filters at certain layers.

\paragraph{Online Pruning}

We firstly empirically investigate the effectiveness of the proposed pruning loss. After setting $\lambda_3=0$, we train VGG-19 network by switching off/on respectively (set $\lambda_2=0$ and $\lambda_2=10^{-4}$) the pruning loss on CIFAR-10 dataset. By adding the proposed pruning loss, we observe improved accuracy of 0.9325 compared to 0.3254 at pruning ratio of 70\%.  When pruning at 80\%, the network without pruning loss can not be constructed due to insufficient preserved filters at certain layers, whereas the network trained with pruning loss can attain an accuracy of 0.9298. This experiment demonstrates that the proposed pruning loss enables online pruning which dynamically selects the architectures while evolving filters to achieve extremely compact structures. 

Fig. \ref{fig:scalings} (b) shows the sorted scaling factors of VGG-19 network trained with pruning loss subject to various target pruning ratios on CIFAR-10. We can observe that given a target pruning ratio, our algorithm adaptively adjusts the distribution of scaling factors to accommodate the pruning operation. Such a dynamic evolution warrants little accuracy loss at a considerably high pruning ratio, as opposed to the static offline pruning approaches, \eg network-slimming, where pruning operation is isolated from the training process causing considerable accuracy loss or even network destruction. 

\begin{figure}[t!]
	\centering
	\includegraphics[width=\linewidth]{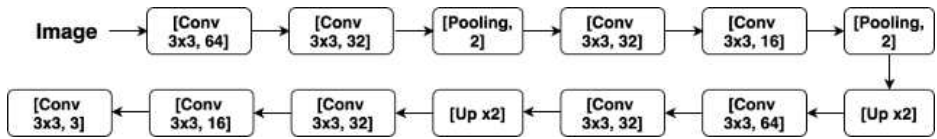}
	\caption{Network architecure for image compression. 
	} \label{fig:comparch}
\end{figure}

\begin{table*}[t!]
	\caption{Results of image compression on CIFAR-100 dataset}
	\centering 
	\begin{tabular}{lccccc} 
		\toprule
		Models & PSNR &  Params &  Pruned (\%)  &  FLOPs  &  Pruned (\%)\\
		\midrule
		Base-line &  30.13 & 75888 & - & 46M & -  \\
		Ours &  29.12  (-3\%)  &  43023 & 43\% & 23M & 50\%  \\
		Ours &  28.89 (-4\%) &  31663 & 58\% & 17M & 63\%  \\
		\bottomrule
	\end{tabular} \label{tbl:compression} 
\end{table*}

\subsection{Image Compression}

The proposed approach is applied on end-to-end image compression task which follows a general autoencoder architecture as illustrated in Fig. \ref{fig:comparch}. We utilize general scaling layer which is added after each convolutional layer, with each scaling factor initialized as 1. The evaluation is performed on CIFAR-100 dataset. We train all the networks from scratch using Adam with mini-batch size 128 for 600 epochs. The learning rate is set to 0.001 and MSE loss is used. The results are listed in Table. \ref{tbl:compression} where both parameters and floating-point operations  (FLOPs) are reported. Our method can save about 40\% - 60\% parameters and 50\% - 60\% computational cost with minor lost of performance (PSNR).

\subsection{Audio Classification}
\begin{figure}[t!]
	\centering
	\includegraphics[width=\linewidth]{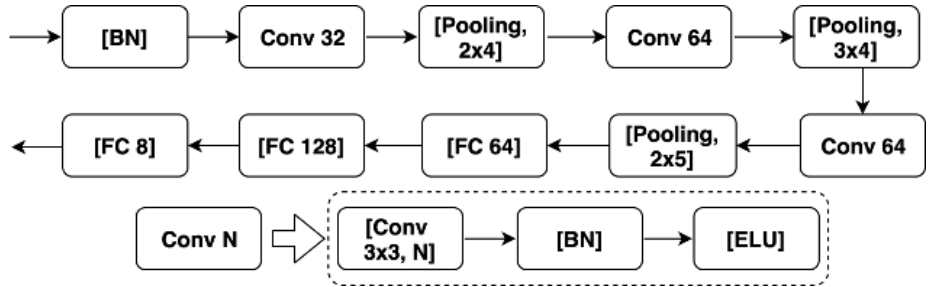}
	\caption{Network architecure for music genre classification. 
	} \label{fig:genrearch}
\end{figure}

We further apply our method in audio classification task, particularly \emph{music genre classification}. The preprocessing of audio data is similar with \cite{lidy2016parallel} and produces Mel spectrogram matrix of size 80$\times$80. The network architecture is illutrated in Fig. \ref{fig:genrearch}, where the scaling layer is added after both convolutional layers and fully connected layers. The evaluation is performed on ISMIR Genre dataset. We train all the networks from scratch using Adam with mini-batch size 64 for 50 epochs. The learning rate is set to 0.003. The results are listed in Table. \ref{tbl:genere} where both parameters and FLOPs are reported. Our approach saves about 92\% parameters while achieves 1\% higher accuracy, saving 80\% computational cost. With a minor loss of about 1\%, 99.5\% parameters are pruned, resulting in an extreme narrow network with $\times$50 times speedup.

\begin{table*}[t!]
	\caption{Results of music genre classification on ISMIR Genre dataset}
	\centering 
	\begin{tabular}{lccccc} 
		\toprule
		Models & Accuracy &  Params &  Pruned (\%)  &  FLOPs  &  Pruned (\%)\\
		\midrule
		Base-line & 0.808 & 106506 & - & 20.3M & -  \\
		Ours & 0.818 (+1\%) &  8056 & 92.5 & 4M & 80.3  \\
		Ours & 0.798 (-1.3\%) &  590 & 99.5 & 0.44M & 98.4  \\
		\bottomrule
	\end{tabular} \label{tbl:genere} 
\end{table*}

\section{CONCLUSIONS}

In this paper, we have proposed a novel approach to simultaneously learning architectures and features in deep neural networks. This is mainly underpinned by a novel pruning loss and online pruning strategy which explicitly guide the optimization toward an optimal architecture driven by a target pruning ratio or model size. The proposed pruning loss enabled online pruning which dynamically selected the architectures while evolving filters to achieve extremely compact structures. In order to improve the feature representation power of the remaining filters, we  further proposed to enforce the diversities between filters for more effective feature representation which in turn improved the trade-off between architecture and accuracies. We conducted comprehensive experiments to show that the interplay between architecture and feature optimizations improved the final compressed models in terms of both models sizes and accuracies for various tasks on both visual and audio data.

\bibliographystyle{splncs03}
\bibliography{NN_compress,NeuralNetworkCollection,computerVisionCollection}

\begin{thebibliography}{10}
\providecommand{\url}[1]{\texttt{#1}}
\providecommand{\urlprefix}{URL }

\bibitem{cano2006ismir}
Cano, P., G{\'o}mez~Guti{\'e}rrez, E., Gouyon, F., Herrera~Boyer, P.,
  Koppenberger, M., Ong, B.S., Serra, X., Streich, S., Wack, N.: Ismir 2004
  audio description contest  (2006)

\bibitem{Cheng2017}
{Cheng}, Y., {Wang}, D., {Zhou}, P., {Zhang}, T.: {A Survey of Model
  Compression and Acceleration for Deep Neural Networks}. ArXiv e-prints  (Oct
  2017)

\bibitem{PruningNip_2018arXiv}
{Crowley}, E.J., {Turner}, J., {Storkey}, A., {O'Boyle}, M.: {Pruning neural
  networks: is it time to nip it in the bud?} ArXiv e-prints arXiv:1810.04622
  (Oct 2018)

\bibitem{han2015learning}
Han, S., Pool, J., Tran, J., Dally, W.: Learning both weights and connections
  for efficient neural network. In: Advances in neural information processing
  systems. pp. 1135--1143 (2015)

\bibitem{Harandi2016}
{Harandi}, M., {Fernando}, B.: {Generalized BackPropagation, \'Etude De Cas:
  Orthogonality}. ArXiv e-prints  (Nov 2016)

\bibitem{hassibi1993second}
Hassibi, B., Stork, D.G.: Second order derivatives for network pruning: Optimal
  brain surgeon. In: Advances in neural information processing systems. pp.
  164--171 (1993)

\bibitem{he2015delving}
He, K., Zhang, X., Ren, S., Sun, J.: Delving deep into rectifiers: Surpassing
  human-level performance on imagenet classification. In: Proceedings of the
  IEEE international conference on computer vision. pp. 1026--1034 (2015)

\bibitem{he2016deep}
He, K., Zhang, X., Ren, S., Sun, J.: Deep residual learning for image
  recognition. In: Proceedings of the IEEE conference on computer vision and
  pattern recognition. pp. 770--778 (2016)

\bibitem{DataDrivenSS_Huang2018}
Huang, Z., Wang, N.: Data-driven sparse structure selection for deep neural
  networks. In: ECCV (2018)

\bibitem{krizhevsky2009learning}
Krizhevsky, A., Hinton, G.: Learning multiple layers of features from tiny
  images. Tech. rep., Citeseer (2009)

\bibitem{lecun1990optimal}
LeCun, Y., Denker, J.S., Solla, S.A.: Optimal brain damage. In: Advances in
  neural information processing systems. pp. 598--605 (1990)

\bibitem{Lib}
{Li}, H., {Kadav}, A., {Durdanovic}, I., {Samet}, H., {Graf}, H.P.: {Pruning
  Filters for Efficient ConvNets}. ArXiv e-prints  (Aug 2016)

\bibitem{lidy2016parallel}
Lidy, T., Schindler, A.: Parallel convolutional neural networks for music genre
  and mood classification. MIREX2016  (2016)

\bibitem{NetSlim_Liu2017d}
Liu, Z., Li, J., Shen, Z., Huang, G., Yan, S., Zhang, C.: {Learning Efficient
  Convolutional Networks through Network Slimming}. ICCV pp. 2736--2744 (2017),
  \url{http://arxiv.org/abs/1708.06519}

\bibitem{RethinkPrune_2018arXiv}
{Liu}, Z., {Sun}, M., {Zhou}, T., {Huang}, G., {Darrell}, T.: {Rethinking the
  Value of Network Pruning}. ArXiv e-prints arXiv:1810.05270 (Oct 2018)

\bibitem{DivNet_2015arXiv}
{Mariet}, Z., {Sra}, S.: {Diversity Networks: Neural Network Compression Using
  Determinantal Point Processes}. ArXiv e-prints arXiv:1511.05077 (Nov 2015)

\bibitem{IntreprePrune_Qin2018}
{Qin}, Z., {Yu}, F., {Liu}, C., {Zhao}, L., {Chen}, X.: {Interpretable
  Convolutional Filter Pruning}. ArXiv e-prints  (Oct 2018)

\bibitem{simonyan2014very}
Simonyan, K., Zisserman, A.: Very deep convolutional networks for large-scale
  image recognition. arXiv preprint arXiv:1409.1556  (2014)

\bibitem{sutskever2013importance}
Sutskever, I., Martens, J., Dahl, G., Hinton, G.: On the importance of
  initialization and momentum in deep learning. In: International conference on
  machine learning. pp. 1139--1147 (2013)

\bibitem{Wen2016}
Wen, W., Wu, C., Wang, Y., Chen, Y., Li, H.: {Learning Structured Sparsity in
  Deep Neural Networks}. {NIPS}  521(12),  61--4 (2016)

\bibitem{Yu2011}
Yu, Y., Li, Y.f., Zhou, Z.h.: {Diversity Regularized Machine}. IJCAI pp.
  1603--1608 (2011)

\bibitem{Zhou2016a}
Zhou, H.: {Less is More : Towards Compact CNNs}. ECCV pp. 1--5 (2016)

\end{thebibliography}

\end{document}